\documentclass[a4paper,english,conferences,final]{IEEEtran}
\usepackage[T1]{fontenc}
\usepackage[latin9]{inputenc}
\usepackage{verbatim}
\usepackage{amstext}
\usepackage{amssymb}
\usepackage{graphicx}

\makeatletter

\pdfpageheight\paperheight
\pdfpagewidth\paperwidth

\newcommand{\lyxdot}{.}

\@ifundefined{showcaptionsetup}{}{%
 \PassOptionsToPackage{caption=false}{subfig}}
\usepackage{subfig}
\makeatother

\usepackage{babel}
\begin{document}
\title{Improve Cost Efficiency of Active Learning over Noisy Dataset}
\author{%
\begin{minipage}[t]{0.3\textwidth}%
\begin{center}
Zan-Kai Chong\\
\emph{Independent Researcher}\\
Malaysia\\
zankai@ieee.org
\par\end{center}%
\end{minipage}%
\begin{minipage}[t]{0.3\textwidth}%
\begin{center}
Hiroyuki Ohsaki\\
\emph{School of Science and Technology\\
Kwansei Gakuin University}\\
Japan\\
ohsaki@kwansei.ac.jp
\par\end{center}%
\end{minipage}%
\begin{minipage}[t]{0.3\textwidth}%
\begin{center}
Bryan Ng\\
\emph{
School of Engineering \& Computer Science\\
Victoria University of Wellington}\\
New Zealand
\\
bryan.ng@ecs.vuw.ac.nz
\par\end{center}%
\end{minipage}}

\maketitle
\thispagestyle{empty}
\begin{abstract}
Active learning is a learning strategy whereby the machine learning
algorithm actively identifies and labels data points to optimize its
learning. This strategy is particularly effective in domains where
an abundance of unlabeled data exists, but the cost of labeling these
data points is prohibitively expensive. In this paper, we consider
cases of binary classification, where acquiring a positive instance
incurs a significantly higher cost compared to that of negative instances.
For example, in the financial industry, such as in money-lending businesses,
a defaulted loan constitutes a positive event leading to substantial
financial loss. To address this issue, we propose a shifted normal
distribution sampling function that samples from a wider range than
typical uncertainty sampling. Our simulation underscores that our
proposed sampling function limits both noisy and positive label selection,
delivering between 20\% and 32\% improved cost efficiency over different
test datasets.
\end{abstract}

\section{Introduction}

Active learning is a specialized form of machine learning that aims
to optimize the learning process by intelligently selecting the data
points that are most beneficial for model training \cite{ren2021survey,budd2021survey,settles2009active}.
Unlike traditional supervised learning, which uses all available labeled
examples, active learning focuses using data points where the model
is most uncertain and hence stands to gain the most information. This
approach is especially valuable in scenarios where labeled data is
scarce or expensive to obtain, such as in medical imaging, natural
language processing, or robotics.

Building on this foundational concept, uncertainty sampling emerged
as one of the earliest and most influential techniques in active learning.
Initially popularized by Lewis \cite{lewis1994sequential}, this approach
showed promise in enhancing the performance of text classification
models. The method was later expanded upon by Schohn and Cohn \cite{schohn2000less},
who applied it to Support Vector Machine (SVM) classification, further
solidifying its role as a potent tool for model improvement. Over
the years, uncertainty sampling has been extensively researched and
applied in various domains, featuring prominently in studies by \cite{yang2015multi,zhu2008active,lughofer2017online,yang2016active,wang2017uncertainty}.
While much of the work in this area has been empirical or heuristic,
there has also been significant theoretical exploration, with contributions
from researchers like \cite{mussmann2018uncertainty,raj2022convergence}.

While uncertainty sampling has earned its reputation as a cornerstone
in the active learning landscape, it is by no means the only strategy
in use. Other approaches, such as query-by-committee \cite{seung1992query},
expected model change \cite{settles2007multiple}, expected error
reduction \cite{roy2001toward}, and expected variance reduction \cite{wang2015ambiguity},
offer alternative ways to implement active learning effectively. These
various methodologies further extend the adaptability and applicability
of active learning, ensuring it remains a vibrant field of study.
For a more in-depth understanding of active learning, interested readers
can refer to the survey paper such as \cite{settles2009active,hino2020active},
and the more recent \cite{tharwat2023survey}.

Despite the advantages of sampling efficiency, active learning presents
several challenges. First, there's the cold start issue, wherein the
effectiveness of uncertainty sampling relies heavily on the initial
model's accuracy in capturing the overview of existing data distribution.
A poorly fitted initial model reduces the efficacy of both subsequent
queries and overall model performance. Second, real-world datasets
often contain noise or may have features that are not adequately representative,
making the regions of uncertainty potentially less valuable for querying.
Third, labeling costs can vary between positive and negative instances;
for instance, in the financial services industry, identifying a loan/mortgage
default is substantially more costly than pinpointing a reliable paymaster.
For a more comprehensive discussion on the practical challenges of
active learning, refer to \cite{tharwat2023survey}.

Recent active learning research has made significant advances, particularly
in the field of deep learning. For example, \cite{hacohen2022active}
addresses the cold start issue in deep learning models by establishing
different querying strategies for high and low budget regimes. \cite{younesian2021qactor}
operates under the assumption of an abundance of pre-labeled instances,
which may contain noise. They employ a high-quality model to identify
and remove dubious or noisy labels, turning to an oracle for final
verification and adjustment.%

In this paper, we assume that the dataset contains both labeled and
unlabeled data, with no free weakly-labeled data, unlike what is commonly
found in deep learning-related studies. We assume that overlapping
features contribute to dataset labeling noise and present ambiguity
for the classification model. To address these issues, we introduce
a broader sampling spectrum to prevent the model from over-sampling
noisy regions. Furthermore, we propose a metric for cost efficiency
that considers the distinct labeling costs for positive instances.

The remainder of this paper is organized as follows: Section 2 introduces
the foundational algorithms of random sampling and uncertainty sampling,
which serve as benchmarks for our proposed method. In Section 3, we
elucidate our proposed sampling function through mathematical modeling
and algorithmic description. Section 4 presents simulation results,
comparing random sampling, uncertainty sampling, and our proposed
dsampling function. Finally, Section 5 offers conclusions drawn from
our work.

\section{Random Sampling and Uncertainty Sampling }

This section introduces random sampling and uncertainty sampling,
which serve as benchmarks for the simulation. Then, the general sampling
algorithm is presented.

\subsection{Random Sampling}

Random sampling (or random samples) is a fundamental concept in the
field of machine learning and statistical analysis, where it serves
as a primary data selection method. It operates under the premise
of providing every instance an equal chance of being selected, thereby
mitigating the risk of bias and promoting diversity within the chosen
sample \cite{kearns1994introduction}.

The strength of random sampling lies in its simplicity and representativeness.
As it doesn't rely on any additional information or complicated selection
mechanisms, it is easily implemented, even in scenarios where preliminary
data understanding is limited. Due to the randomness in selection,
the sample often captures the overall distribution of the population
well, contributing to the generalizability of the learned model.

However, random sampling is not without its weaknesses. In situations
where the distribution of labels is imbalanced, or certain classes
of data are less prevalent, random sampling overlooks these under
represented labels. Consequently, the learned model suffers from performance
issues when predicting less common classes \cite{kaur2019systematic}.
Moreover, as random sampling does not consider the informativeness
or relevance of instances, it might lead to the selection of redundant
or non-informative instances, causing a potential waste of computational
resources.

Despite these drawbacks, random sampling remains a common and crucial
sampling strategy in machine learning projects due to its inherent
simplicity and the fair representation it provides. Various studies
have employed random sampling as a baseline or comparative method
in research on active learning, reinforcement learning, and other
machine learning disciplines.

\subsection{Uncertainty Sampling}

Uncertainty sampling is a data selection strategy in active learning,
where instances about which the model is most uncertain are chosen
for labelling. Essentially, it seeks to label instances for which
the current predictive model has the lowest confidence, thereby focusing
on the most informative samples to improve the model's learning efficiency
\cite{lewis1994sequential}.

One of the principal advantages of uncertainty sampling lies in its
potential to maximize the learning outcomes from a limited pool of
labeled instances. By focusing on samples that the model finds hard
to classify, it aids in refining the decision boundary and enhancing
overall model performance, leading to more efficient use of computational
resources compared to random sampling.

Despite these advantages, uncertainty sampling has its limitations.
Primarily, it runs the risk of gravitating towards complex or noisy
instances, as these are the samples where the model often experiences
the most uncertainty. This bias can lead to model overfitting, as
the learning may focus excessively on outliers or irregular instances.

\subsection{General Sampling Algorithm\label{subsec:General-Sampling-Algorithm}}

The abovementioned random sampling and uncertainty sampling can be
elaborated with the following general algorithm.
\begin{itemize}
\item Step 1 - Initialization: Start with an initial classification model
that has been trained on a labeled dataset (known data pool, $L$.
Additionally, prepare an unlabeled dataset (unknown data pool, $U$).
Define stopping criteria such as a predetermined number of queries,
denoted by $N$. Each query is indexed by and integer $q$, where
$q\leq N$
\item Step 2 - Instance Selection: In this step, select instances from $U$
using either random sampling or uncertainty sampling, based on the
strategy chosen for the simulation. For random sampling, randomly
select $N$ instances from $U$ without consideration of the model's
output. For uncertainty sampling, use the existing model to predict
the class probabilities for each instance in $U$. Select instances
based on measures of uncertainty, for example least confidence, margin
sampling, or entropy. After querying their true labels, move the $N$
selected instances from $U$ to $L$. 
\item Step 3 - Model Refinement: Using the updated labeled dataset $L$,
retrain the classification model. If the stopping criteria are not
met, return to Step 2 and continue the process.
\end{itemize}

\section{Sampling over Uncertainty and Noise Regions}

The assumptions and the mathematical model of the proposed sampling
function are stated and discussed in the following sections.

\subsection{Assumptions and Setup}

As mentioned above, we use two distinct pools of data: one labeled
and another unlabeled. The initial size of the labeled data pool is
modest, while the unlabeled data pool is significantly larger. We
also assume that the distribution of the overall population remains
constant throughout the querying process. Our study focuses on a binary
classification problem with labels categorized as 0 and 1 or True
or False.

It is important to note that the simulations involve assessing the
actual model performance, using the mean values of the Area Under
the Curve (AUC) and the F1 score, derived from \emph{three} distinct
test data sets. These performance outcomes are for referencing purposes
only and are available solely to the observers during the simulation.
They do not in any way influence the results of the different algorithms
tested.

Finally, we assume that the active learning process concludes after
a predetermined number of queries.

\subsection{Shifted Normal Sampling: Inspired by the Uncertainty Region and Noisy
Region}

Suppose we have the final model of an active learning process and
its associated prediction probability, denoted as $p_{i}^{\text{final}}$,
for a binary classification dataset. Here, $i$ is the index of the
unlabeled instances and it may be omitted for brevity.

\begin{figure}
\begin{centering}
\includegraphics[width=0.95\columnwidth]{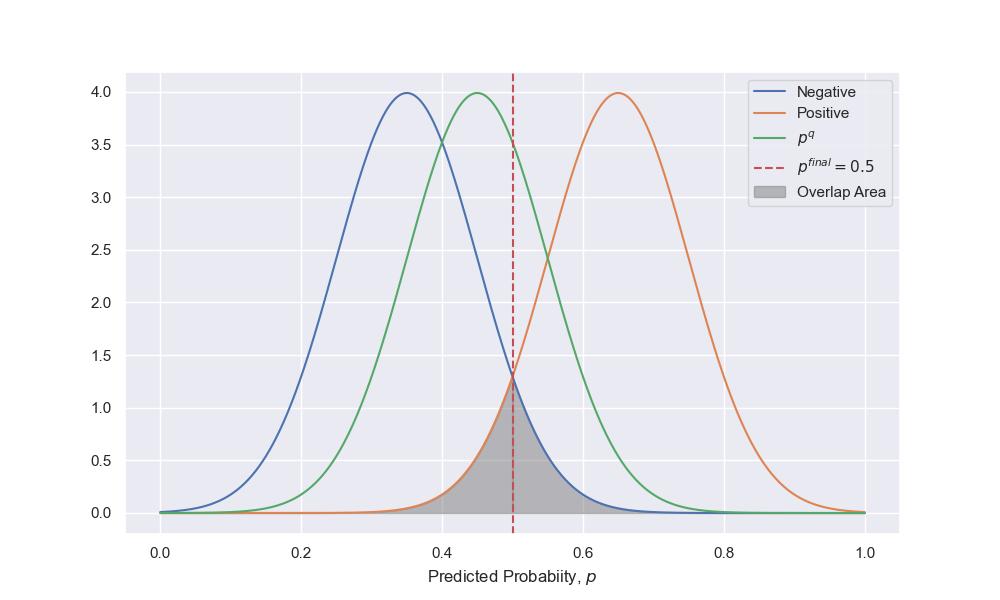}
\par\end{centering}
\caption{Optimal probability distributions for positive and negative events,
exhibiting minimal overlap area, and shifted normal samping function.
\label{fig:small-overlap}}
\end{figure}

By using $p^{\text{final}}$, we map all unlabelled instances to the
range $[0,1]\in\mathbb{R}$, where $\mathbb{{R}}$is the set of real
numbers. Within this range, instances with high predicted probablility
are unambiguously associated with their respective labels of 0 and
1. In contrast, those around $p^{_{\text{final}}}=0.5$ oscillate
between the labels as the query progresses. This ambiguity is attributed
to their location in an overlap region, which we term as \emph{noise}
here, as illustrated in Fig. \ref{fig:small-overlap}. Let the integer
$q$ denote the $q$-th query to the set $U$ (see Section \ref{subsec:Sampling-Algorithm}).
Then the region where $p^{q}\approx0.5$ serves as the uncertainty
region for interim models to query. In contrast, the area around $p^{\text{final}}=0.5$
is considered the noise region, as it is uncertainly classified the
final model.

Now, assume that we are using uncertainty sampling in querying instances
of predicted probabilities $p^{q}$ close to the uncertainty region,
i.e., in the range of $(0.5-\delta,0.5+\delta)$, for a small value
$\delta$ at $q$-th query, where $\delta\in\mathbb{R}+$ and $\delta$
closes to 0 . Here, the $p^{q=1}=0.5$ from the initial model might
significantly differ from $p^{\text{final}}=0.5$. Yet, as queries
continue, we anticipate a gradual refinement in the interim models,
drawing instances closer and closer to $p_{\text{final}}=0.5$. Let
$\varphi(q)$ be the actual sampling distribution of the interim model
with reference to the final model, i.e., 
\begin{equation}
\varphi(q)=\left\{ p_{i}^{\text{final}}:\forall i,p_{i}^{q}\in\left[0.5-\delta,0.5+\delta\right]\right\} .
\end{equation}
 We expect $\varphi(q)$ to follow a normal distribution, represented
as $\varphi\sim\text{N}(0.5,\sigma)$, where $\sigma$ shrinks as
$q$ increases.

The essence of uncertainty sampling is to query within the uncertainty
region, which overlaps with the noise region as queries progress.
While initially minimal, this overlap can increase as interim models
are refined, further pushing $\varphi(q)$ towards the noise region.
We argue that narrow sampling spectrum over the noise region is the
cause of performance fluctuations in the interim models. To address
this issue, rather than sampling within the range $[0.5-\delta,0.5+\delta]$,
we opt for a wider normal distribution, denoted as $\text{N}(0.5,\sigma)$.
This allows us to shift the selection of instances closer to $p^{q}=0.5$,
while controlling the sampling spread via the standard deviation $\sigma$.

The details of the algorithm will be elaborated in Section \ref{subsec:Sampling-Algorithm}.

\subsection{Imbalance Labelling Cost}

We assume that the model performance metric such as AUC, F1, etc,
is measured and denoted as $\lambda(q)$, where higher $\lambda(q)$
represents better performance and $0\leq\lambda(q)\leq1;\lambda(q)\in\mathbb{{R}}$,
for the $q$-th query.  . To tune the performance measurement of
the sampling algorithm in terms of its responsiveness to varying proportions
of positive instances, we introduce a metric designated as \emph{cost
efficiency}, i.e.,

\begin{equation}
\eta(q)=\frac{\lambda(q)}{\zeta(q)\,C},
\end{equation}
where $\eta\in\mathbb{{R}}$ , $\eta\geq0$ and $\zeta(q):\zeta(q)\geq0$
represent the proportion of positive instances within the known dataset.

Here, $C$ is the cost associated with a single positive instance
relative to a negative instance and $C\geq1$. When the costs for
both positive and negative instances are identical, $C$ is 1. Conversely,
if the cost of a positive instance is three times the cost of a negative
instance, $C$ is 3. For the ensuing simulation, we adopt $C=1$ as
our standard value. Overall, a higher value of $\eta$ corresponds
to increased efficiency.

As mentioned earlier, we assume that the labeling costs for positive
events are significantly higher than those for negative events in
this case. Consequently, we adjust the normal sampling distribution
slightly to the left, i.e., $\text{N}(0.45,\sigma)$, such that the
sampling is slightly biased towards negative events (Fig. \ref{fig:small-overlap}).
This adjustment will yield relatively fewer positive labels in $\zeta(q)$,
thereby improving $\eta(q)$.

This metric will be used in simulation results in Section \ref{sec:Simulation-Results}.

\subsection{Normal Sampling Algorithm \label{subsec:Sampling-Algorithm}\label{subsec:Normal-Sampling-Algorithm}}

We illustrate the algorithm of the proposed sampling function as follows.

\subsubsection{Step 1 - Initialization}

Commence with a pre-existing classification model, accompanied by
a labelled dataset (the known data pool, $L$),  and an unlabelled
dataset (the unknown data pool, $U$). Define a set of stopping criteria,
such as a predefined number of queries, denoted as $N$.

\subsubsection{Step 2 - Instance Selection}

This step is analogous to uncertainty sampling, wherein instances
are selected based on predicted probabilities from all instances in
$U$. However, diverging from uncertainty sampling, instances are
selected with predicted probabilities in accordance with a normal
distribution. In our simulation, we will use a beta function, i.e.,
$\text{Beta}\left(\alpha,\beta\right)$ to approximate the normal
distribution, where $\alpha$ and $\beta$ are adjusted such that
peak values occur at $p=0.45$. Correspondingly, most instances are
sampled near to $p=0.45$, without disregarding instances from other
ranges. Then, queried instances will be incorporated into $L$.

\subsubsection{Step 3 - Model Refinement}

A new model is built using $L$. If the stopping criteria have not
been met, the process will be repeated from Step 2.

\section{Simulation Results\label{sec:Simulation-Results}}

This section investigates the performance of random sampling, uncertainty
sampling and shifted normal sampling through simulation. The discussion
commences with a description of the simulation setup, followed by
an overview of the artificial datasets that we used. Finally, the
simulation results are delineated.

\subsection{Simulation Overview}

In this paper, we created synthetic classification datasets using
Scikit-learn \cite{scikit-learn}. The datasets comprise four feature
columns, and each dataset has different levels of noise. Due to page
limits, we present results for two sets of noise parameters in our
simulations. The results shown are representative of our extensive
simulation over different simulation parameters.

Our primary focus was on binary classification, assuming a large data
pool with an equal number of positive and negative events. For the
purpose of our study, we divided the datasets randomly into three
sections. The first, a labeled data pool, contained 10 instances.
The second, an unknown data pool, held 1000 instances. Lastly, we
had three test data pools, each containing 1000 instances. These test
data were used to evaluate the performance of our model after each
query. The parameter \emph{class\_sep} regulates the separation between
the label classes. A smaller value reduces the separation and makes
classification harder.

In line with the approach outlined in Section \ref{subsec:Normal-Sampling-Algorithm},
we adapted the sampling algorithm during Step 2 to meet our simulation
objective, choosing from random sampling, uncertainty sampling, or
shifted normal sampling. Each query necessitated the selection of
two instances from $U$ to be annotated, revealing their true labels.
This process was repeated across a total of 20 queries. Unless otherwise
specified, all the presented graphs are the results of 30 rounds of
simulations and are plotted with 99\% confidence intervals (CI).

The models were built using the Generalized Linear Model (GLM) from
H2O \cite{H2OAutoML20}, set to the optimal parameters and with no
further feature engineering. It is important to note that the ability
of GLM to accurately identify positive events can vary depending on
the characteristics of the dataset.

\subsection{Results}

\begin{figure}
\begin{centering}
\subfloat[]{\begin{centering}
\includegraphics[width=0.7\columnwidth]{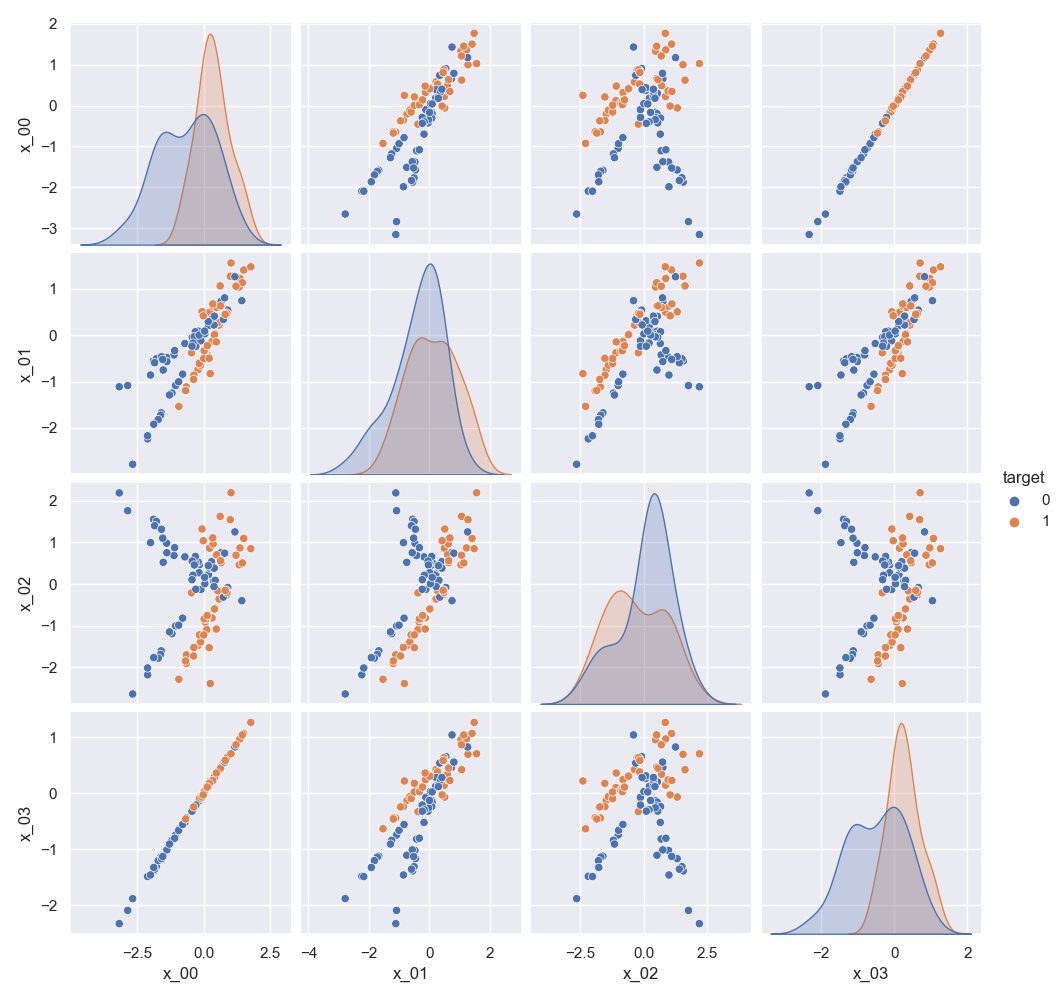}
\par\end{centering}
}
\par\end{centering}
\begin{centering}
\subfloat[]{\begin{centering}
\includegraphics[width=0.7\columnwidth]{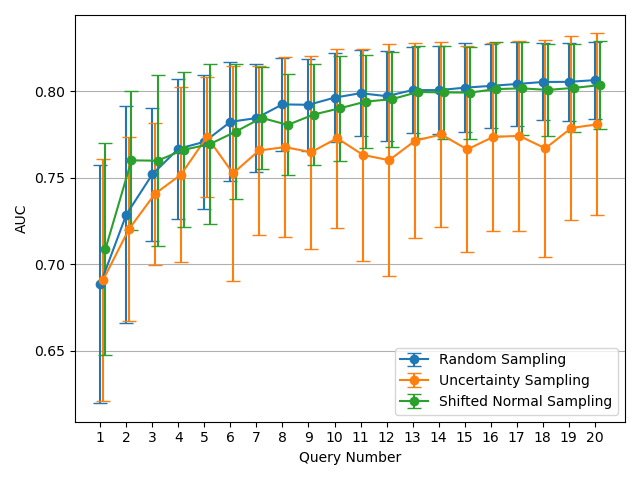}
\par\end{centering}
}
\par\end{centering}
\begin{centering}
\subfloat[]{\begin{centering}
\includegraphics[width=0.7\columnwidth]{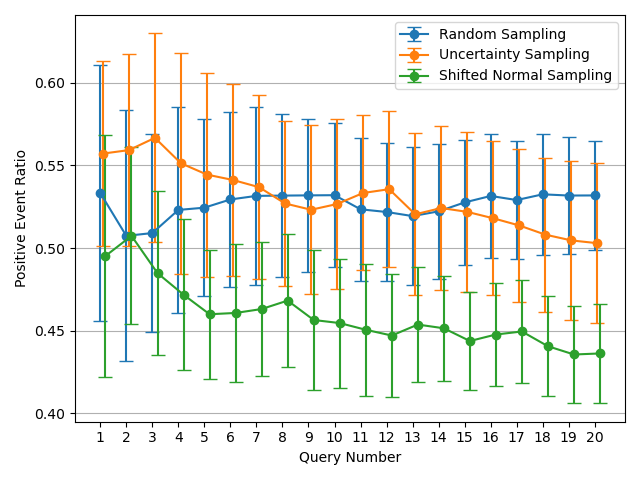}
\par\end{centering}
}
\par\end{centering}
\begin{centering}
\subfloat[]{\begin{centering}
\includegraphics[width=0.7\columnwidth]{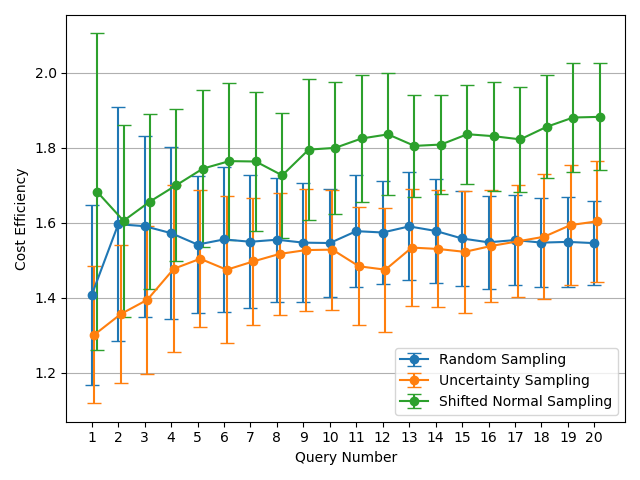}
\par\end{centering}
}
\par\end{centering}
\caption{Illustration of the classification dataset with parameter \emph{class\_sep}=0.5
indicating overlap dataset. The sub-figures represent (a) bivariate
distribution, (b) model performance measured in AUC, (c) ratio of
positive events, and (d) cost efficiency across each query. \label{fig:classification-cs0.5}}
\end{figure}

Fig. \ref{fig:classification-cs0.5} shows the performance for a classification
dataset, set with the parameter \emph{class\_sep}=0.5 for overlap
dataset, where its bivariate pairwise distribution is illustrated
in Fig. \ref{fig:classification-cs0.5}(a). Fig. \ref{fig:classification-cs0.5}(b)
charts the AUC, revealing an increasing trend as the number of queries
increases with better refined interim models. In particular, we note
that shifted normal sampling and random sampling exhibit similar AUC
values, while uncertainty sampling underperforms on average. Then,
Fig. \ref{fig:classification-cs0.5}(c) sheds light on their corresponding
positive event ratio at 99\% CI. We observe both random sampling and
uncertainty sampling stay around $\zeta^{\left(\text{random}\right)}=0.53$
and $\zeta^{\left(\text{uncertainty}\right)}=0.50$ whereas shifted
normal sampling converge approximately at $\zeta^{\left(\text{normal}\right)}=0.43$
at $q=20$. It is worth noting that although shifted normal sampling
has similar AUC performance as random sampling, the former has lower
positive event ratio comparatively, results in better cost efficiency,
i.e., achieving $\eta^{\left(\text{normal}\right)}=1.88$ for shifted
normal sampling --- about 20\% improvement as compared to uncertainty
sampling $\eta^{\left(\text{uncertainty}\right)}=1.60$ and random
sampling $\eta^{\left(\text{random}\right)}=1.54$ at final query.

\begin{figure}
\begin{centering}
\subfloat[]{\begin{centering}
\includegraphics[width=0.7\columnwidth]{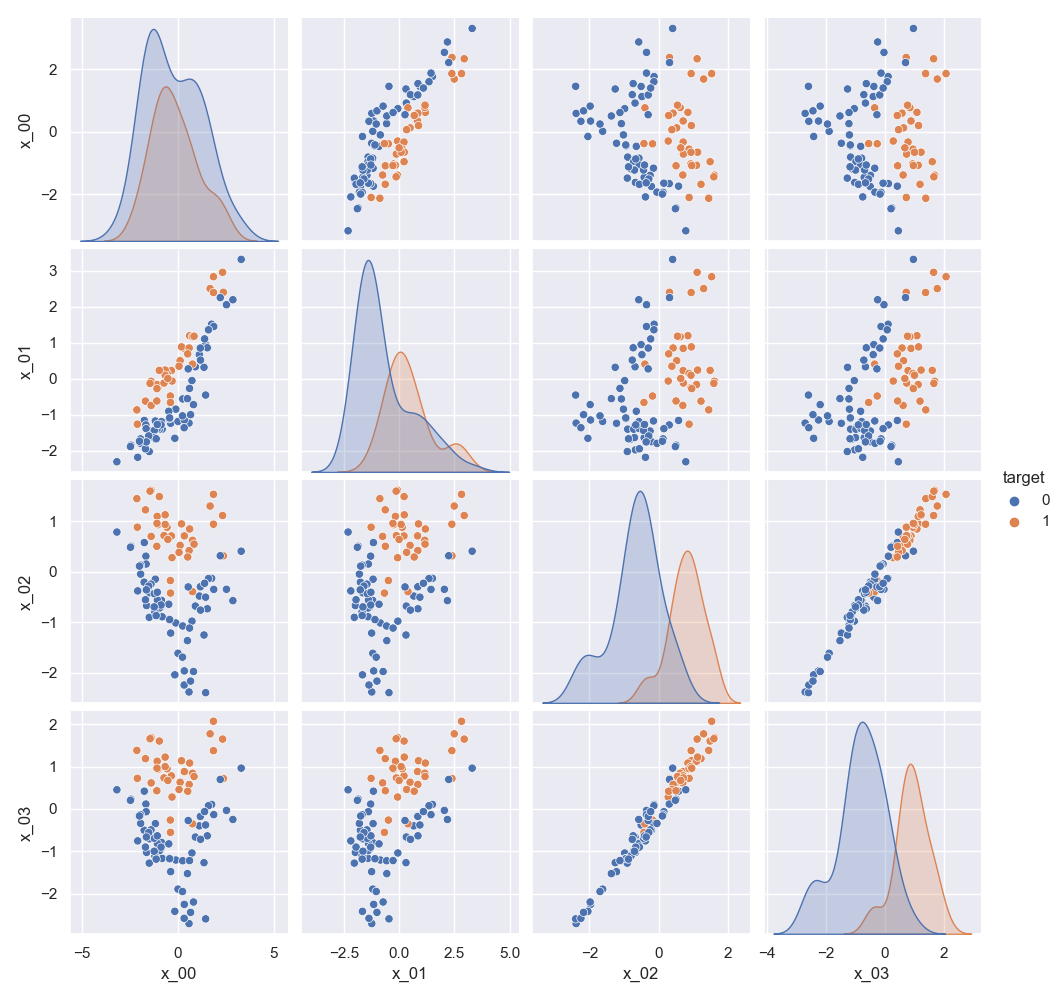}
\par\end{centering}
}
\par\end{centering}
\begin{centering}
\subfloat[]{\begin{centering}
\includegraphics[width=0.7\columnwidth]{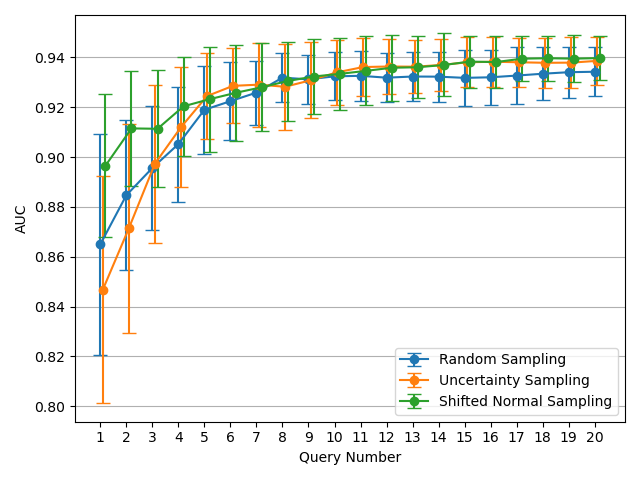}
\par\end{centering}
}
\par\end{centering}
\begin{centering}
\subfloat[]{\begin{centering}
\includegraphics[width=0.7\columnwidth]{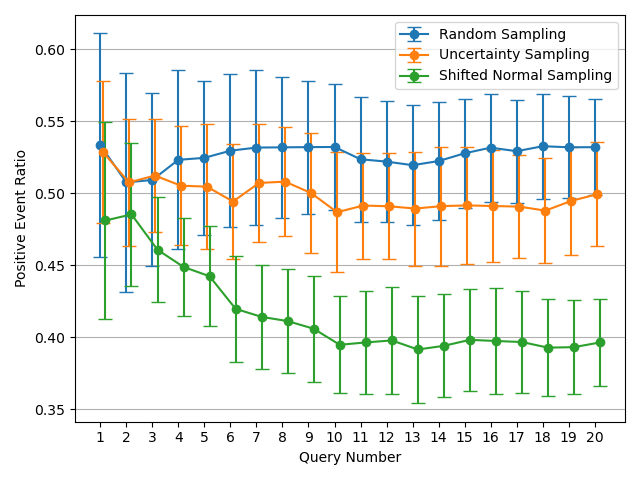}
\par\end{centering}
}
\par\end{centering}
\begin{centering}
\subfloat[]{\begin{centering}
\includegraphics[width=0.7\columnwidth]{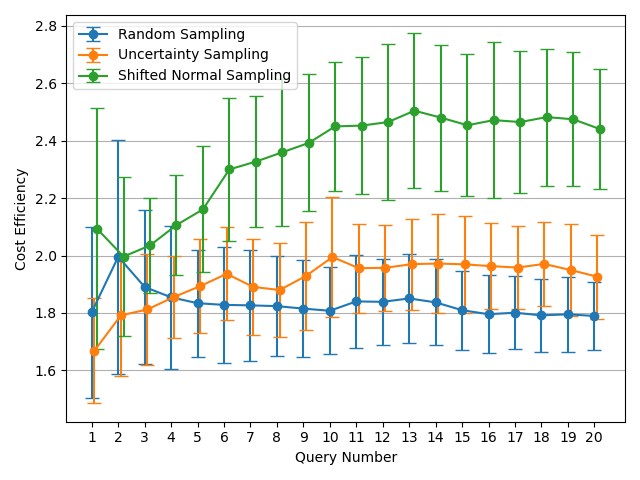}
\par\end{centering}
}
\par\end{centering}
\caption{Representation of the classification dataset for parameter \emph{class\_sep}=1.0.
(a) bivariate distribution, (b) AUC, (c) positive event ratio and
(d) cost efficiency.\label{fig:classification-cs1.0}}
\end{figure}

Simulations on classification datasets with \emph{class\_sep}=1.0
show minor overlap, as in Fig. \ref{fig:classification-cs1.0}. Generally,
all three sampling functions exhibits competitive AUC performance
with the their respective mean positive event ratios all within the
99\% CI. Since shifted normal sampling has the lowest positive event
ratio, the correponding cost efficiency stays highest, i.e. $\eta^{\left(\text{normal}\right)}=2.44$,
about 32\% improvement as compared to $\eta^{\left(\text{random}\right)}=1.78$
and $\eta^{\left(\text{uncertainty}\right)}=1.92$ at final query.

\section{Conclusion}

This study aimed to examine the effectiveness of various sampling
algorithms---namely random sampling, uncertainty sampling, and the
proposed shifted normal sampling, in the context of noisy binary classification
datasets. We considered cost efficiency, particularly in scenarios
where labelling positive instances is more costly than negative instances.
Our findings reveal that shifted normal sampling strikes a robust
balance between AUC performance and improves cost efficiency up to
32\%.

\bibliographystyle{IEEEtran}
\bibliography{zankai}

\end{document}